# DETECTION OF FAKE FACES IN VIDEOS

[1]Mathias Russel Rudolf Richard, [2]M Shamanth, [3]Dr. Vijayalakshmi M N
[1,2]Students, [3]Associate Professor
[1]Master of Computer Applications,
[1]RV College of Engineering®, Bengaluru, India.

*Abstract :* Deep learning methodologies have been used to create applications that can cause threats to privacy, democracy and national security and could be used to further amplify malicious activities. One of those deep learning-powered applications in recent times is synthesized videos of famous personalities. According to Forbes, Generative Adversarial Networks(GANs) generated fake videos growing exponentially every year and the organization known as Deeptrace had estimated an increase of deepfakes by 84% from the year 2018 to 2019. They are used to generate and modify human faces, where most of the existing fake videos are of prurient non-consensual nature, of which its estimates to be around 96% and some carried out impersonating personalities for cyber crime. In this paper, available video datasets are identified and a pretrained model BlazeFace is used to detect faces, and a ResNet and Xception ensembled architectured neural network trained on the dataset to achieve the goal of detection of fake faces in videos. The model is optimized over a loss value and log loss values and evaluated over its F1 score. Over a sample of data, it is observed that focal loss provides better accuracy, F1 score and loss as the gamma of the focal loss becomes a hyper parameter. This provides a k-folded accuracy of around 91% at its peak in a training cycle with the real world accuracy subjected to change over time as the model decays.

*Keywords*-Generative Adversarial Networks, BlazeFace, ResNet, Xception, Deepfake

## I. INTRODUCTION

Advancements in Deep Learning techniques are being used to create softwares which are proving to be a major threat to national security, privacy and the democracy of a country. One of those deep learning-powered applications in recent times is synthesized media of famous personalities. According to Forbes, the word deepfake became famous on the Internet in late 2017. Deepfake is just a combination of two words "deep learning" and "fake". They are just modified media in which a person's face is modified with someone else's. This new innovation is obtained from a deep learning method known as Generative Adversarial Networks (GANs). These modified or fake videos are rising at an alarming rate every year.[1] According to a report from startup Deeptrace, in the beginning of 2019 there were only about 7,964 deepfake videos online, but that figure just rose to 14,678 nine months later. This figure is still growing and there is no doubt it's going to rise exponentially.

As indicated by a report in The Wall Street Journal, in March 2019, a CEO of an anonymous U.K-based energy firm was scammed using the technologies of synthesized media. He was scammed for an amount of $243,000[2] when he followed the orders to promptly move €220,000 (approx. $243,000) to a Hungarian provider thinking the call to be from his chief, the CEO of association's the German parent organization

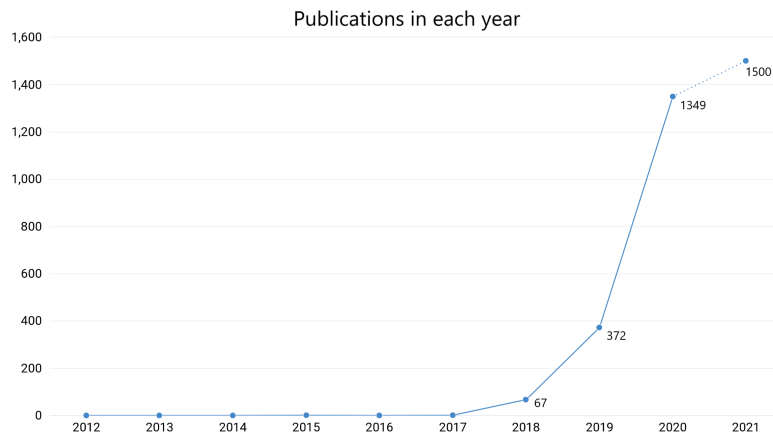

Figure 1: Publication of papers related to synthetic media from 2012 to 2021, sourced from https://app.dimensions.ai on December 17, 2021.

The fake faces in videos are generated using a method known as GANs, which were a revelation in the field of deep learning and creation of synthetic data. It is used to generate fake faces from videos in such a way that the target person will act or say things the source person does. A deepfake video is created by swapping one person's face and replacing it with another, usually a deep learning algorithm called variational auto-encoder is used to create such fake faced videos. There are various categories of AI synthesized content nowadays, but mainly these synthetic media can fall into either lip-sync or puppet-master. The former refers to synthesized videos which are modified to create mouth variations aligning with sound recordings and the latter refers to synthesized videos where a target person called the puppet is made to follow the facial variations such as movements of eye and movement of head like another person called as the master.

## II. LITERATURE SURVEY

The authors Ian J. Goodfellow and et al [3] in their paper Generative Adversarial Nets depicted the models given a train set, how it creates new data with similar distributions of information as the training set. It proposes a framework for estimating generative models via an adversarial process, in which one simultaneously trains two models.

The authors Shruti Agarwal, Hao Li and et al [4] in their paper Protecting World Leaders Against Deep Fakes clarifies the methodology for identifying impersonators of leaders around the world. It describes a forensic technique that models facial expressions and movements that typify an individual's speaking pattern.

The authors Joseph Redmon, Santosh Divvala, Ali Farhadi and et al [5] in their paper You Only Look Once: Unified, Real-Time Object Detection clarifies the Single-Shot utilized for Object Detection. It frames object detection as a regression problem to spatially separated bounding boxes and associated class probabilities.

The authors Darius Afchar, Isao Echizen and et al [6] in their paper MesoNet: a Compact Facial Video Forgery Detection Network explains one of the techniques to consequently and proficiently recognize face altering in recordings or videos. It presents a method to efficiently detect face tampering in videos, and particularly focuses on two recent techniques used to generate hyper-realistic forged videos.

The authors Tadas Baltrusaitis, Peter Robinson and et al [7] in their paper Cross-dataset learning and person-specific normalisation for automatic Action Unit detection clarifies a Facial Action Unit assessment and event discovery framework which are dependent on appearance. It demonstrates the benefits of using a simple and efficient median based feature normalisation technique that accounts for person specific neutral expressions.

The authors Robert Chesney and Danielle Keats Citron [8] in their paper Deep fakes: A looming challenge for privacy, democracy, and national security gives an appraisal of the causes and outcomes of fake media prompting troublesome change, and furthermore investigates the current and possible devices for reacting to it. It aims to provide the first in-depth assessment of the causes and consequences of this disruptive technological change concerning deepfakes and to explore the existing and potential tools for responding to it.

The author Francois Chollet [9] in their paper Xception: Deep learning with separable convolutions clarifies the convolutional neural net architecture similar to Inception which marginally beats Inception V3 on the ImageNet dataset. It provides an interpretation of Inception modules in convolutional neural networks as being an intermediate step in-between regular convolution and the depthwise separable convolution operation.

The authors Zhou, Chen and Wei [10] in their paper Multi-attentional deepfake detection audits an attention based neural network or a transformer based neural network architecture to detect deepfakes. The paper proposes a new multi-attentional deepfake detection network. Specifically, it consists of three key components: 1) multiple spatial attention heads to make the network attend to different local parts; 2) textural feature enhancement block to zoom in the subtle artifacts in shallow features; 3) aggregate the low-level textural feature and high-level semantic features guided by the attention maps.

The authors Agarwal, Varshney and et al [11] in their paper Limits of Deepfake Detection: A Robust Estimation Viewpoint audits a statistical methodology stripped from any deep learning processing or usage of neural networks for trying to detect deepfakes. A robust statistics view of GANs is considered to bound the error probability for various GAN implementations in terms of their performance.

The authors Falko Matern, Marc Stamminger and et al [12] in their paper Exploiting Visual Artifacts to Expose Deepfakes and Face Manipulations depicted the current facial altering techniques and several characteristic artifacts from their processing pipelines. It also shows that relatively simple visual artifacts can be already quite effective in exposing such manipulations, including Deepfakes and Face2Face.

The authors David Guera and Edward J Delp [13] in their paper Deepfake Video Detection Using Recurrent Neural Networks clarifies Recurrent neural nets (RNN) that figures out how to characterize whether a video has been tampered. This paper evaluates the method against a large set of deepfake videos collected from multiple video websites.

The authors Yuezun Li, Siwei Lyu and et al [14] in their paper In ictu oculi: Exposing artificial intelligence made fake video recordings by distinguishing eye blinks clarifies different techniques utilizing the identification of eye blinking in the video recordings, which is a physiological sign that isn't well presented in the combined fake video recordings. The method is tested over benchmarks of eye-blinking detection datasets and also shows promising performance on detecting videos generated with DeepFake.

## III. IMPLEMENTATION

BlazeFace, the pretrained face detection model is used to detect faces in the extracted frames. It is a machine learning model developed by Google to rapidly detect the location and key points of faces.[15] The next step is to select suitable Neural Networks based on the performance which may either be ResNet, XceptionNet or any other neural network. After this a Neural Network model is trained using the chosen neural network algorithm in Google Cloud Platform's AI Platform service. This model will be hypertuned by changing the parameters and the desired model is obtained. This model is tested against the test video data available to get the accuracy. After the final model is obtained, It could be deployed to production and over the course of time using improved models.

The ResNet model utilized has 48 convolutional layers and 2 fully connected dense layers, the Xception model architecture has 36 convolutional layers which are residually connected with the layers in front of each layer. Here the convolutional layers are depthwise separable convolution layers which makes an assumption that the features are independent of each other and which allows for faster training and better performance than standard convolutional layers.

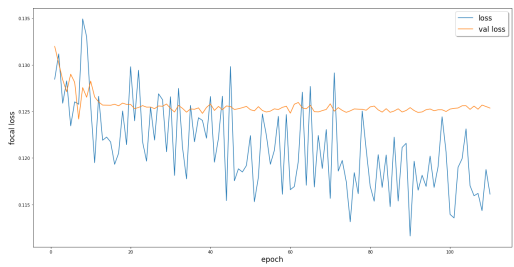

Figure 2: Train and validation loss on a 100 epochs with 5 frames as input and batch size as 32 with Focal loss as its loss function

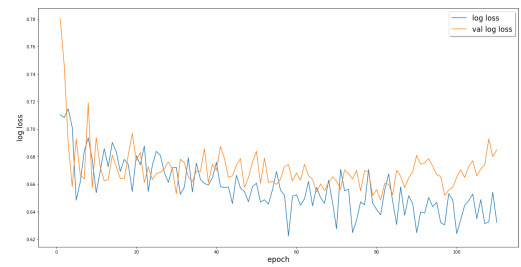

Figure 3: Train and validation log loss on a 100 epochs with 5 frames as input and batch size as 32 with Focal loss as its loss function

Figure 2 shows the loss function values over a 100 epochs with 5 frames as its inputs in the input layer with focal loss as its loss function. It shows a plateauing of the validation loss after a set of 40 epochs, which indicates that the model could use either a different set of parameters or changes in architecture.

Figure 3 shows the log loss function values over a 100 epochs with 5 frames as its inputs in the input layer with focal loss as its loss function. It shows a plateauing of the validation loss after a set of 40 epochs encompassing information from the loss function, one can conclude that there are negative effects on the generalization of the model as it trains over more than 80.

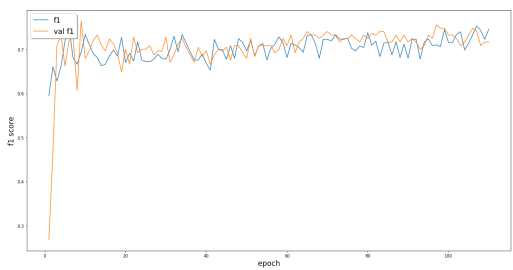

Figure 4: Train and validation F1 scores on a 100 epochs with 5 frames as input and batch size as 32 with Focal loss as its loss function

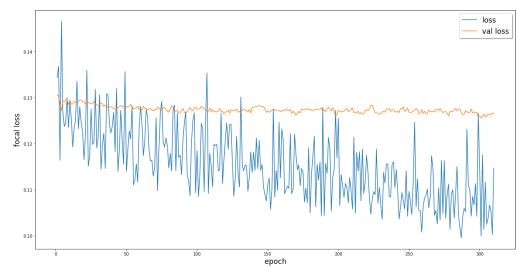

Figure 5: train and validation loss on a 300 epochs with 6 frames as input and batch size as 64 with Focal loss as its loss function

Figure 4 shows the F1 score of the model trained over a 100 epochs with 5 frames. The validation F1 and training F1 are over close to each other indicating that there's room for improvement.

Figure 5 shows the loss function values over 300 epochs with 6 frames and a batch size of 64 as its inputs in the input layer with focal loss as its loss function. There's a clear sign of overfitting, one could reduce the number of epochs trained or take the best validation loss values and save that model over 300 epochs or decrease the batch size.

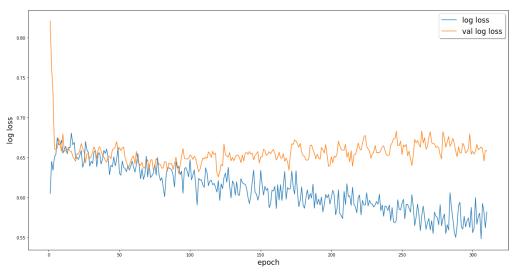

Figure 6: Train, validation log loss on a 300 epochs with 6 frames as input and batch size as 64 with Focal loss as its loss function

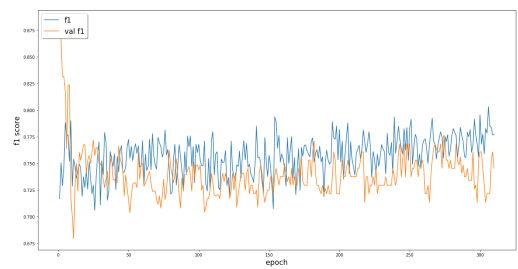

Figure 7: Train, validation F1 score on 300 epochs with 6 frames as input and batch size as 64 with Focal loss.

Figure 6 shows the log loss function values over 300 epochs with 6 frames as its inputs in the input layers with focal loss as its loss function. The validation log loss and loss function are highly correlated and they seem to be deviating from each other which is as expected another sign of overfitting which in this case is seens in a linear trend.

Figure 7 shows the F1 score of the model trained over 300 epochs with 6 frames. The validation F1 and training F1 are more volatile than the F1 score indicated by the model trained over a 100 epochs , 32 batch size and 5 frames. This volatility could be caused by not having the gradient reach a global minima and might need a more fine tuned learning rate.

## IV. ANALYSIS OF DEEPFAKE DETECTION

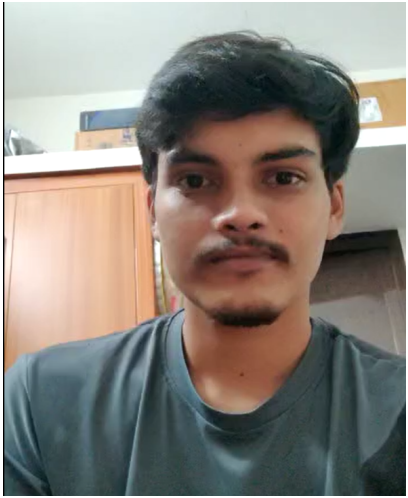 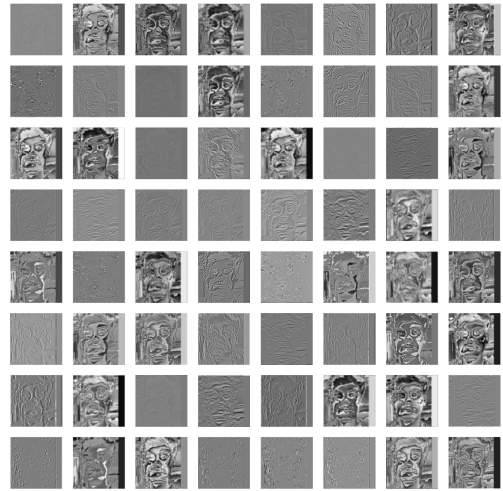

Figure 8: Input Raw image         Figure 9: Layer 1

Figure 8 shows one of the frames taken from a test input video which would be filtered through the learnt layers of one of the 2 model architectures namely ResNet50.
Figure 9 shows the first convolutional layers visual representation of the features that are being extracted from each of the filters where each filter has extracted a unique set of features.

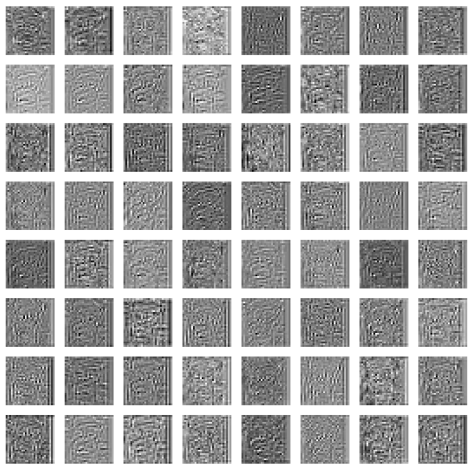 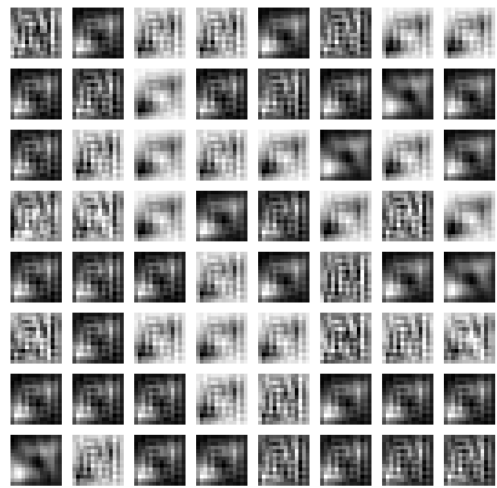

Figure 10: Layer 24         Figure 11: Layer 48

Figure 10 shows a more abstract representation of the features that are passed from the previous layer namely from the layer 23, which are then forward passed to layer 24 as shown. Figure 11 shows the features that the model's last convolutional layer has learnt and which will then be passed to the next layer which are 2 sets of fully connected dense layers to make predictions about the certainty of the video being a deepfake.

## V. RESULTS AND CONCLUSION

In this paper, available video datasets are identified. This dataset contains videos, images and its associated labels. The study involves BlazeFace to detect faces and a ResNet architectured neural network trained on the media to achieve the goal of detection of fake faces in videos. Modern techniques used to create such fake faces can similarly be used to detect and recognise it. On this dataset, a deep neural network is trained and detects if the entity is real or fake and tests its accuracy on a validation set separate to test the accuracy. This model is deployed as a service on cloud platforms which the general public can utilize, allowing for the utilization of such tools more widespread due to increase in its access.

Once the whole project is developed, the model is incorporated and the working project appears to the users. Once these progressions are consolidated, the upkeep period of the project life cycle begins. This includes staying up with the latest with innovation in the field of deep learning algorithms, scaling it by including more features and making the prediction more effective in accuracy.

The upsides of this kind of a process is, it advances collaboration, incremental improvement, simple to oversee, adaptable sort of programming improvement, usefulness can be developed quickly and illustrated. In the end the gain in efficiency and accuracy to detect fake faces in videos is beneficial to the general public.